\documentclass[10pt,journal,letterpaper,compsoc]{IEEEtran}
\usepackage[cmex10]{amsmath}
\usepackage{amsfonts}
\usepackage{amssymb}
\usepackage{bm}
\usepackage{graphicx}
\usepackage{graphics}
\usepackage[applemac]{inputenc}
\usepackage{cite}
\usepackage{mdwtab}
\usepackage{subfigure}
\usepackage{color}
\usepackage{xspace}
\usepackage{url}
\usepackage{tabularx}
\usepackage{epstopdf}
\usepackage{multirow}

\usepackage{caption}

\newcolumntype{C}[1]{>{\centering}m{#1}}

\makeatletter
\DeclareRobustCommand\onedot{\futurelet\@let@token\@onedot}
\def\@onedot{\ifx\@let@token.\else.\null\fi\xspace}

\def\eg{\emph{e.g}\onedot} 
\def\ie{\emph{i.e}\onedot} 
\def\cf{\emph{cf}\onedot}

\def\etal{\emph{et al}\onedot}
\makeatother

\newcommand{\Fig}{Fig.\xspace}

\newcommand{\Sec}{Sec.\xspace}
\newcommand{\Tab}{Tab.\xspace}

\newcolumntype{L}[1]{>{\raggedright\let\newline\\\arraybackslash\hspace{0pt}}m{#1}}
\newcolumntype{C}[1]{>{\centering\let\newline\\\arraybackslash\hspace{0pt}}m{#1}}
\newcolumntype{R}[1]{>{\raggedleft\let\newline\\\arraybackslash\hspace{0pt}}m{#1}}

\newcommand{\MOTChallenge}{{\it MOTChallenge}\xspace}
\newcommand{\MOTsix}{{\it MOT16}\xspace}
\newcommand{\MOTfive}{{\it MOT15}\xspace}
\newcommand{\MOTseven}{{\it MOT17}\xspace}

\newcommand{\CVPR}{{\it CVPR19}\xspace}
\newcommand{\MOTtwenty}{{\it MOT20}\xspace}

\newcommand{\dismeas}{d}
\newcommand{\simthresh}{t_d}  		

\graphicspath{{figures/},{figures/moving/},{figures/static/},{figures/annotations/}}

\definecolor{darkgreen}{rgb}{0,.75,0}
\definecolor{gray40}{gray}{.40}

   \teaser{
   \centering
   \hspace{0.005\linewidth}
   \includegraphics[width=.92\linewidth]{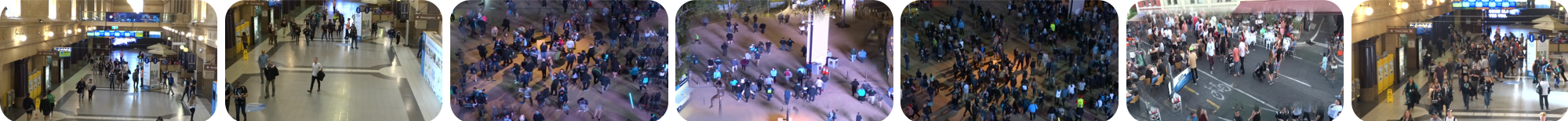}
  }

\begin{document}
\title{MOT20: A benchmark for multi object tracking in crowded scenes}

\author{Patrick Dendorfer, 
        Hamid Rezatofighi,
        Anton Milan,
        Javen Shi,
            Daniel Cremers, 
        Ian Reid, \\
        Stefan Roth,
        Konrad Schindler, 
        and Laura Leal-Taix{\'e}
\thanks{P.~ Dendorfer and L.~Leal-Taix{\'e} are with the Dynamic Vision and Learning Group at TUM Munich, Germany.}
\thanks{H.~Rezatofighi, J.~Shi, and I.~Reid are with the Australian Institute for Machine Learning and the School of Computer Science at University of Adelaide.}
\thanks{A.~Milan is with Amazon, Berlin, Germany. This work was done prior to joining Amazon.} 
\thanks{K.~Schindler is with the Photogrammetry and Remote Sensing Group at ETH Zurich, Switzerland.}
\thanks{D.~Cremers is with the Computer Vision Group at TUM Munich, Germany.}
\thanks{S.~Roth is with the Department of Computer Science, Technische Universit{\"a}t Darmstadt, Germany.}
\thanks{Primary contacts: patrick.dendorfer@tum.de, leal.taixe@tum.de }
}

\IEEEcompsoctitleabstractindextext{%

\begin{abstract}

Standardized benchmarks are crucial for the majority of computer vision applications.
Although leaderboards and ranking tables should not be over-claimed, benchmarks often
provide the most objective measure of performance and are therefore important guides 
for research.
The benchmark for Multiple Object Tracking, \MOTChallenge, was launched with the goal to establish a standardized evaluation of multiple object tracking methods. The challenge focuses on multiple people tracking, since pedestrians are well studied in the tracking community, and precise tracking and detection has high practical relevance. Since the first release, \MOTfive~\cite{MOTChallenge:arxiv:2015}, \MOTsix~\cite{MOTChallenge:arxiv:2016}, and \MOTseven~\cite{MOTChallenge:arxiv:2016} have tremendously contributed to the community by introducing a clean dataset and precise framework to benchmark multi-object trackers. 
In this paper, we present our \MOTtwenty
benchmark, consisting of 8 new sequences depicting very crowded challenging scenes. The benchmark was presented first at the 4$^{\textnormal{th}}$ BMTT MOT Challenge Workshop at the Computer Vision and Pattern Recognition Conference (CVPR) 2019, and gives to chance to evaluate state-of-the-art methods for multiple object tracking when handling extremely crowded scenarios.

\end{abstract}

}

\maketitle

\IEEEdisplaynotcompsoctitleabstractindextext

\IEEEpeerreviewmaketitle


\section{Introduction}
\label{sec:introduction}
Since its first release in 2014, \MOTChallenge has attracted more than $1,000$ active users who have successfully submitted their trackers and detectors to five different challenges, spanning $44$ sequences with $2.7M$ bounding boxes over a total length of $36k$ seconds. 
As evaluating and comparing multi-target tracking methods is not trivial (\emph{cf.~e.g.}~\cite{Milan:2013:CVPRWS}), \MOTChallenge provides carefully annotated datasets and clear metrics to evaluate the performance of tracking algorithms and pedestrian detectors. Parallel to the \MOTChallenge all-year challenges, we organize workshop challenges on multi-object tracking for which we often introduce new data.

\begin{figure*}

 \includegraphics[width=\linewidth]{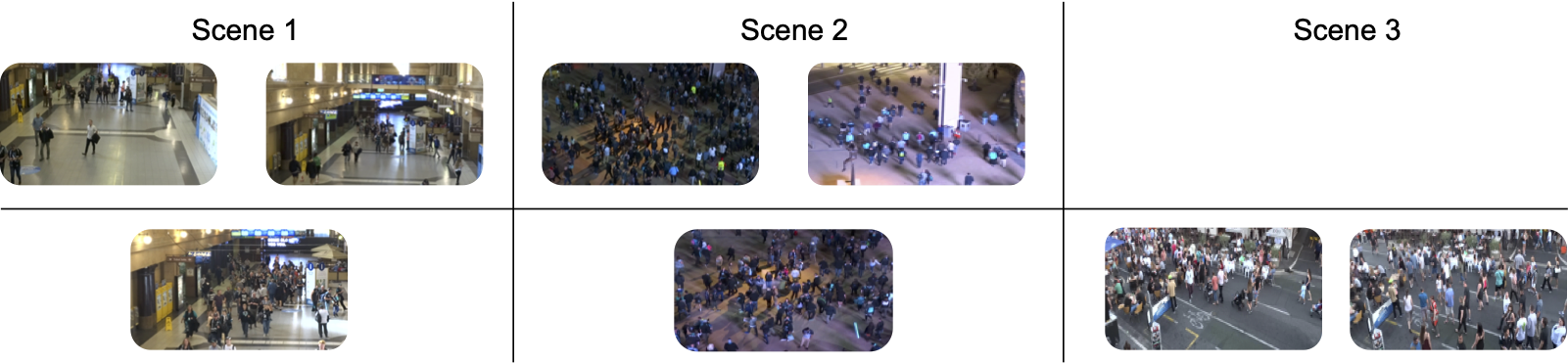}
 \caption{An overview of the \MOTtwenty dataset. The dataset consists of 8 different sequences from 3 different scenes. The test dataset has two known and one unknown scene. Top: \textbf{training sequences}; bottom: \textbf{test sequences}. }
 \label{fig:dataOverview}
\end{figure*}

In this paper, we introduce the  \MOTtwenty benchmark, consisting of $8$ novel sequences out of $3$  very crowded scenes. All sequences have been carefully selected and annotated according to the evaluation protocol of previous challenges~\cite{{MOTChallenge:arxiv:2015}, MOTChallenge:arxiv:2016}. This benchmark addresses the challenge of very crowded scenes in which the density can reach values of $246$ pedestrians per frame. The sequences were filmed in both indoor and outdoor locations, and include day and night time shots. Figure~\ref{fig:dataOverview} shows the split of the sequences of the three scenes into training and testing sets. The testing data consists of sequences from known as well as from an  unknown scenes in order to measure the genralization capabilities of both detectors and trackers. We make available the images for all sequences, the ground truth annotations for the training set as well as a set of public detections (obtained from a Faster R-CNN trained on the training data) for the tracking challenge.

The \MOTtwenty challenges and all data, current ranking and submission guidelines can be found at:
\begin{center}
\url{ht tp://www.motchallenge.net/}
\end{center}

\begin{figure*}
\centering
\begin{minipage}{0.33\textwidth}
\centering
 \includegraphics[width = 0.8\textwidth]{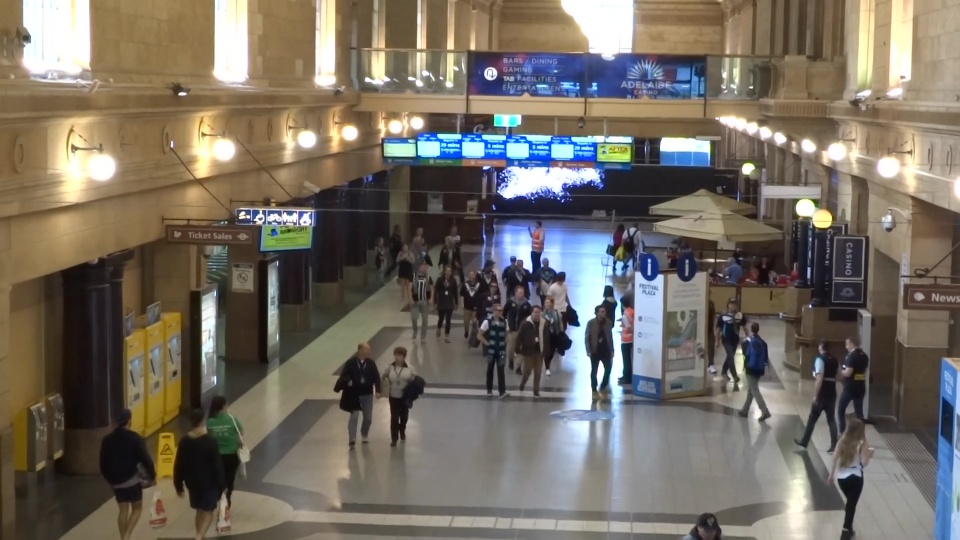}
\end{minipage}%
\begin{minipage}{0.33\textwidth}
\centering 
 \includegraphics[width = 0.8\textwidth]{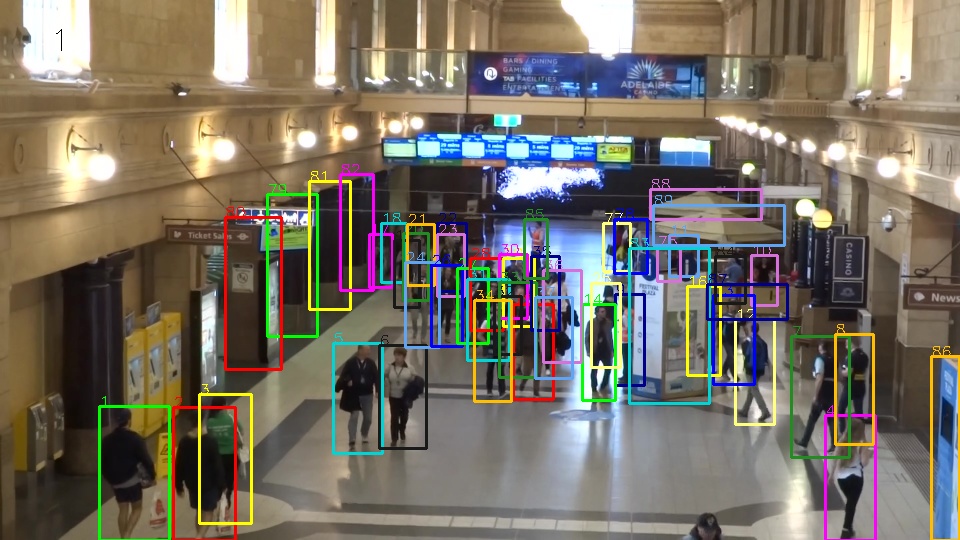}
\end{minipage}%
\begin{minipage}{0.33\textwidth}
\centering
 \includegraphics[width = 0.8\textwidth]{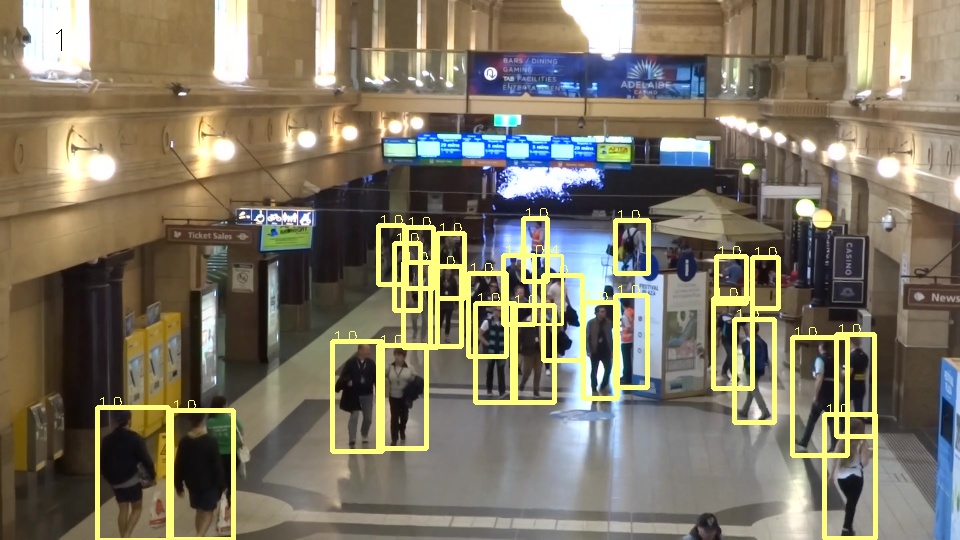}
\end{minipage}%

 \caption{we provide for the challenges. Left: Image of each frame of the sequences; middle: ground truth labels including all classes. Only provided for training set; right: public detections from trained Faster R-CNN. }
 \label{fig:providedData}
\end{figure*}

\section{Annotation rules}
\label{sec:anno-rules}

For the annotation of the dataset, we follow the protocol introduced in \MOTsix, ensuring that every moving person or vehicle within each sequence is annotated
with a bounding box as accurately as possible. In the following, we 
define a clear protocol that was obeyed throughout the entire dataset to 
guarantee consistency.

\subsection{Target class}
In this benchmark, we are interested in tracking moving objects in videos. In particular, we are interested in evaluating multiple people tracking algorithms, hence, people will be the center of attention of our annotations. 
We divide the pertinent classes into three categories: \\
(i) {\it moving} pedestrians;\\
 (ii) people that are {\it not in an upright position}, not moving, or artificial representations of humans;  and \\
 (iii) {\it vehicles} and {\it occluders}.
 
In the first group, we annotate all moving pedestrians that appear in the field of view and can be determined as such by the viewer.  Furthermore, if a person \emph{briefly} bends over or squats, \eg, to pick something up or to talk to a child, they shall remain in the standard \emph{pedestrian} class.
The algorithms that submit to our benchmark are expected to track these targets.

In the second group, we include all people-like objects whose exact classification is ambiguous and can vary depending on the viewer, the application at hand, or other factors. We annotate all static people, \eg, sitting, lying down, or do stand still at the same place over the whole sequence. 
The idea is to use these annotations in the evaluation such that an algorithm is neither penalized nor rewarded for tracking, \eg, a sitting  or not moving person.

In the third group, we annotate all moving vehicles such as cars, bicycles, motorbikes and non-motorized vehicles (\eg, strollers), as well as other potential occluders. These annotations will not play any role in the evaluation, but are provided to the users both for training purposes and for computing the level of occlusion of pedestrians. Static vehicles (parked cars, bicycles) are not annotated as long as they do not occlude any pedestrians.

\section{Datasets}
\label{sec:datasets}

The dataset for the new benchmark has been carefully selected to challenge trackers and detectors on extremely crowded scenes. In contrast to previous  challenges, some of the new sequences show a pedestrian density of $246$ pedestrians per frame. In \Fig~\ref{fig:dataOverview} and \Tab~\ref{tab:dataoverview}, we show an overview of the sequences 
included in the benchmark.

\begin{table*}[tbh]
\begin {center}
 \begin{tabular}{|l| c| c| r| r| r| r| c | c| }
 \hline
 \multicolumn{9}{|c|}{\bf Training sequences} \\ 
 \hline 
      Name & FPS & Resolution & Length & Tracks & Boxes & Density & Description & Source\\ 
      \hline
MOT20-01 & 25      & 1920x1080  & 429 (00:17)  &   74 &      19,870 &      46.32    & indoor &  new    \\
MOT20-02 & 25      & 1920x1080  & 2,782 (01:51) &  270 &     154,742 &      55.62  & indoor   & new    \\
MOT20-03 & 25      & 1173x880   & 2,405 (01:36) &  702 &     313,658 &     130.42 & outdoor, night& new    \\
MOT20-05 & 25      & 1654x1080  & 3,315 (02:13) & 1169 &     646,344 &     194.98   & outdoor, night                                         & new    \\
         
      \hline
      \multicolumn{3}{|c|}{\bf Total training} & {\bf 8,931 (05:57)} & {\bf 2,215} & {\bf 1,134,614} & {\bf 127.04} & &   \\
    \hline
    \multicolumn{9}{c}{\vspace{1em}} \\
    \hline
   \multicolumn{9}{|c|}{\bf Testing sequences} \\ 
 \hline 
      Name & FPS & Resolution & Length & Tracks & Boxes & Density & Description & Source\\ 
      \hline
MOT20-04 & 25      & 1545x1080  & 2,080 (01:23)  & 669 &     274,084 &     131.77 & outdoor, night   & new    \\
MOT20-06 & 25      & 1920x734  & 1,008 (00:40) & 271 &     132,757 &     131.70  &outdoor, day	 & new    \\
MOT20-07 & 25      & 1920x1080	& 585 (00:23) & 111 &      33,101 &      56.58   & indoor	 & new    \\
MOT20-08 & 25      & 1920x734  & 806 (00:32) &  191 &      77,484 &      96.13  & outdoor, day     & new    \\

      \hline
      \multicolumn{3}{|c|}{\bf Total training} & {\bf 4,479  (02:58)} & {\bf 1,242	} & {\bf 517,426} & {\bf 115.52	} & &   \\
      \hline 
    \end{tabular}
  \end{center}
    \caption{Overview of the sequences currently included in the \MOTtwenty benchmark considering pedestrians.}
\label{tab:dataoverview}
\end{table*}

\begin{table*}[tbh]
\begin {center}
\begin{tabular}{|l|r|r|r|r|r|r|}

 \hline
 \multicolumn{7}{|c|}{\bf Annotation classes} \\ 
 \hline 
Sequence & Pedestrian & Non motorized vehicle & Static person 
 & Occluder on the ground & crowd     & Total \\
 \hline
MOT20-01 & 19,870  & 0    & 2,574   & 4,203  & 0   & 26,647  \\
MOT20-02 & 154,742 & 4,021 & 11,128  & 32,324 & 0   & 202,215 \\
MOT20-03 & 313,658 & 1,436 & 22,310 & 16,835 & 2,489 & 356,728 \\
MOT20-04 & 274,084 & 3,110 & 92,251 & 2,080  & 0   & 371,525 \\
MOT20-05 & 646,344 & 7,513 & 90,843  & 6,630  & 0   & 751,330 \\
MOT20-06 & 132,757 & 1,248 & 60,102  & 12,428  & 1,008   & 207,543 \\
MOT20-07 & 33,101  & 800  & 3,685   & 3,510  & 0   & 41,096  \\
MOT20-08 & 77,484 & 4,237 & 52,998  & 9,776  & 806  & 145,301 \\
\hline
Total     & 1,652,040 & 22,365 & 335,891 & 87,786 & 4,303 & 2,102,385\\
      \hline 
    \end{tabular}
  \end{center}
    \caption{Overview of the types of annotations currently found in the \MOTtwenty benchmark.}
\label{tab:dataclasses}
\end{table*}

\subsection{MOT 20 sequences}

We have compiled a total of 8 sequences, of which we use half for 
training and half for testing. The annotations of the testing sequences 
will not be released in order to avoid (over)fitting of the methods to the 
specific sequences. The sequences are filmed in three different scenes. 
Several sequences are filmed per scene and distributed in the train and test sets. One of the scenes though, is reserved for test time, in order to challenge the generalization capabilities of the methods.

The new data contains circa 3 times more bounding boxes for training and testing compared to \MOTseven. All sequences are filmed in high resolution from an elevated viewpoint, and the mean crowd density reaches 246 pedestrians per frame which $10$ times denser when compared to the first benchmark release. Hence, we expect the new sequences to be more challenging for the tracking community and to push the models to their limits when it comes to handling extremely crowded scenes.
In \Tab~\ref{tab:dataoverview}, we give an overview of the training and testing sequence characteristics for the challenge, including the number of bounding boxes annotated.

Aside from pedestrians, the annotations also include other classes like vehicles or bicycles, as detailed in \Sec~\ref{sec:anno-rules}. In \Tab~\ref{tab:dataclasses}, we detail the types of annotations that can be found in each sequence of \MOTtwenty.

\subsection{Detections}
\label{sec:detections}

We trained a Faster R-CNN~\cite{fasterRCNN} with ResNet101~\cite{resnet} backbone on the \MOTtwenty training sequences, obtaining the detection results presented in Table~\ref{tab:detresultsoverview}. This evaluation follows the standard protocol for the \MOTtwenty challenge and only accounts for pedestrians. Static persons and other classes are not considered and filtered out from both, the detections, as well as the ground truth. 

A detailed breakdown of detection bounding boxes on individual sequences is provided in \Tab~\ref{tab:det-performance}.

\begin{table}[hbt]

\begin {center}
 \begin{tabular}{| l |r r r r  |}
  \hline 
  \bf Seq & \bf nDet. & \bf nDet./fr. &\bf min height& \bf max height \\ 
          \hline 
 MOT20-01 &     12610&     29.39&               30&     289 \\
  MOT20-02  &   89837&     32.29&           25&       463 \\
  MOT20-03   & 177347&     73.74&             16&       161 \\
  MOT20-04&    228298&    109.76&              23&       241 \\
  MOT20-05&    381349&    115.04&              26&       245 \\
  MOT20-06&     69467&     68.92&             27&       304 \\
  MOT20-07&     20330&     34.75&           21&       381 \\
  MOT20-08&     43703&     54.22&              37&       302 \\
  
\hline
   Total&   1022941&     76.28&             16    &   463\\      
 \hline 
    \end{tabular}
  \end{center}
    \caption{Detection bounding box statistics.}
\label{tab:det-performance}
\end{table}

For the tracking challenge, we provide these public detections as a baseline to be used for training and testing of the trackers. 
For the \MOTtwenty challenge, we will only accept results on public detections. When later the benchmark will be open for continuous submissions, we will accept both public as well as private detections.

\subsection{Data format}
\label{sec:data-format}

All images were converted to JPEG and named sequentially to a 6-digit file name (\eg~000001.jpg). Detection and annotation files are simple comma-separated value (CSV) files. Each line represents one object instance and contains 9 values as shown in \Tab~\ref{tab:dataformat}.

The first number indicates in which frame the object appears, while
the second number identifies that object as belonging to a trajectory
by assigning a unique ID (set to $-1$ in a detection file, as no ID is
assigned yet). Each object can be assigned to only one trajectory.
The next four numbers indicate the position of the bounding box of the
pedestrian in 2D image coordinates. The position is indicated by the
top-left corner as well as width and height of the bounding box.
This is followed by a single number, which in case of detections
denotes their confidence score.
The last two numbers for detection files are ignored (set to -1).

\begin{table*}[hbt]
\begin {center}
 \begin{tabular}{| c | c| p{13cm}|}
 \hline
     \bf Position & \bf Name & \bf Description\\ 
          \hline 
    1 & Frame number & Indicate at which frame the object is present\\
 2 & Identity number & Each pedestrian trajectory is identified by a
 unique ID ($-1$ for detections)\\
 3 & Bounding box left &  Coordinate of the top-left corner of the pedestrian bounding box\\
  4 & Bounding box top & Coordinate of the top-left corner of the pedestrian bounding box \\
 5 & Bounding box width & Width in pixels of the pedestrian bounding box\\
 6 & Bounding box height & Height in pixels of the pedestrian bounding box\\
 7 & Confidence score & DET: Indicates how confident the detector is that this instance is a pedestrian. \hspace{5cm} GT: It acts as a flag whether the entry is to be considered (1) or ignored (0).  \\
  8 &  Class & GT: Indicates the type of object annotated  \\
 9 & Visibility & GT: Visibility ratio, a number between 0 and 1 that says how much of that object is visible. Can be due to occlusion and due to image border cropping. \\
      \hline 
    \end{tabular}
  \end{center}
    \caption{Data format for the input and output files, both for detection (DET) and annotation/ground truth (GT) files.}
\label{tab:dataformat}
\end{table*}

\begin{table}[hbt]
\begin {center}
 \begin{tabular}{| l | c|}
  \hline 
  \bf Label & \bf ID\\ 
          \hline 
Pedestrian & 1 \\
Person on vehicle & 2 \\
Car & 3 \\
Bicycle & 4 \\
Motorbike & 5 \\
Non motorized vehicle & 6 \\
Static person & 7 \\
Distractor & 8 \\
Occluder & 9 \\
Occluder on the ground & 10 \\
Occluder full & 11 \\
Reflection & 12 \\
Crowd & 13 \\
 \hline 
    \end{tabular}
  \end{center}
    \caption{Label classes present in the annotation files and ID appearing in the 8$^\text{th}$ column of the files as described in \Tab~\ref{tab:dataformat}.}
\label{tab:labelclass}
\end{table}

 An example of such a 2D detection file is:
 \begin{samepage}
\begin{center}
\begin{footnotesize}
  \texttt{1, -1, 794.2, 47.5, 71.2, 174.8, 67.5, -1, -1}\nopagebreak\\
  \texttt{1, -1, 164.1, 19.6, 66.5, 163.2, 29.4, -1, -1}\nopagebreak\\
  \texttt{1, -1, 875.4, 39.9, 25.3, 145.0, 19.6, -1, -1}\nopagebreak\\
  \texttt{2, -1, 781.7, 25.1, 69.2, 170.2, 58.1, -1, -1}\nopagebreak\\
\end{footnotesize}
\end{center}
\end{samepage}

For the ground truth and results files, the 7$^\text{th}$ value (confidence score) acts as a flag whether the entry is to be considered. A value of 0 means that this particular instance is ignored in the evaluation, while a value of 1 is used to mark it as active. 
The 8$^\text{th}$ number indicates the type of object annotated, following the convention of \Tab~\ref{tab:labelclass}. The last number shows the visibility ratio of each bounding box. This can be due to occlusion by another static or moving object, or due to image border cropping.

 An example of such an annotation 2D file is:
 \begin{samepage}
\begin{center}
\begin{footnotesize}
  \texttt{1, 1, 794.2, 47.5, 71.2, 174.8,  1,  1, 0.8}\nopagebreak\\
  \texttt{1, 2, 164.1, 19.6, 66.5, 163.2,  1,  1, 0.5}\nopagebreak\\
  \texttt{2, 4, 781.7, 25.1, 69.2, 170.2, 0, 12, 1.}\nopagebreak\\
\end{footnotesize}
\end{center}
\end{samepage}

In this case, there are 2 pedestrians in the first frame of the sequence, with identity tags 1, 2. 
In the second frame, we can see a static person (class 7), which is to be considered by the evaluation script and will neither count as a false negative, nor as a true positive, independent of whether it is correctly recovered or not.
Note that all values including the bounding box are 1-based, \ie the top left corner corresponds to $(1,1)$.

To obtain a valid result for the entire benchmark, a separate CSV file following the format described above must be created for each sequence and called \texttt{``Sequence-Name.txt''}. All files must be compressed into a single ZIP file that can then be uploaded to be evaluated.

\begin{table*}[tbh]
\begin {center}
 \begin{tabular}{|l| rrrrrrrrrr| }
 \hline
 \multicolumn{11}{|c|}{\bf Training sequences} \\ 
 \hline

      Sequence & AP & Rcll & Prcn &FAR & GT & TP & FP & FN & MODA & MODP \\ 
      \hline
MOT20-01&	0.82&	86.5&	99.5&	0.14&	12945&	11199&	58&	1746&	86.06&	91.61 \\
MOT20-02&	0.82&	85.9&	99.5&	0.15&	93107&	79971&	421&	13136&	85.44&	92.13 \\
MOT20-03&	0.54&	59.0&	98.4&	1.10&	278148&	163988&	2653&	114160&	58.00&	86.00 \\
MOT20-05&	0.63&	64.2&	99.4&	0.60&	528037&	338826&	1979&	189211&	63.79&	87.59\\

    \hline
    \multicolumn{11}{c}{\vspace{1em}} \\
  \hline
   \multicolumn{11}{|c|}{\bf Testing sequences} \\ 
 \hline 
      Sequence & AP & Rcll & Prcn &FAR & GT & TP & FP & FN & MODA & MODP \\ 
      \hline

 MOT20-04&	0.63&	69.7&	98.0&	1.55&	230729&	160783&	3230&	69946&	68.29&	81.41\\
MOT20-06&	0.43&	57.9&	74.4&	12.64&	63889&	37002&	12745&	26887&	37.97&	73.67\\
MOT20-07&	0.78&	83.6&	92.5&	1.89&	16298&	13627&	1106&	2671&	76.83&	79.11\\
MOT20-08&	0.38&	55.2&	61.6&	13.93&	32608&	17998&	11230&	14610&	20.76&	71.55\\

      \hline 
    \end{tabular}
  \end{center}
    \caption{Overview of performance of Faster R-CNN detector trained on the \MOTtwenty training dataset. }
\label{tab:detresultsoverview}
\end{table*}

\section{Evaluation}
\label{sec:evaluation}
Our framework is a platform for fair comparison of state-of-the-art
tracking methods. By providing authors with standardized ground truth
data, evaluation metrics and scripts, as well as a set of precomputed detections, all methods are compared under the exact same conditions, thereby isolating the performance of the tracker from everything else.
In the following paragraphs, we detail the set of evaluation metrics that we provide in our benchmark.

\subsection{Evaluation metrics}
\label{sec:evaluation-metrics}
In the past, a large number of metrics for quantitative evaluation of 
multiple target tracking have been proposed \cite{Smith:2005:CVPRW, 
Stiefelhagen:2006:CLE, Bernardin:2008:CLE, Schuhmacher:2008:ACM, 
Wu:2006:CVPR, Li:2009:CVPR}. Choosing ``the right'' one is largely 
application dependent and the quest for a unique, general evaluation metric
is still ongoing. On the one hand, it is desirable to summarize the performance 
into one single number to enable a direct comparison. On the other hand,
one might not want to lose information about the individual errors made
by the algorithms and provide several performance estimates, which 
precludes a clear ranking.

Following a recent trend \cite{Milan:2014:PAMI, Bae:2014:CVPR, 
Wen:2014:CVPR}, we employ two sets of measures that have established 
themselves in the literature: The \emph{CLEAR} metrics proposed by 
Stiefelhagen \etal \cite{Stiefelhagen:2006:CLE}, and a set of track 
quality measures introduced by Wu and Nevatia \cite{Wu:2006:CVPR}.
The evaluation scripts used in our benchmark are publicly 
available.\footnote{\url{http://motchallenge.net/devkit}}

\subsubsection{Tracker-to-target assignment}
\label{sec:tracker-assignment}
There are two common prerequisites for quantifying the performance of a 
tracker. One is to determine for each hypothesized output, whether it is a 
true positive (TP) that describes an actual (annotated) target, or 
whether the output is a false alarm (or false positive, FP). This 
decision is typically made by thresholding based on a defined distance 
(or dissimilarity) measure $\dismeas$ (see 
\Sec~\ref{sec:distance-measure}). A target that is missed by any 
hypothesis is a false negative (FN). A good result is expected to have 
as few FPs and FNs as possible. Next to the absolute numbers, we also 
show the false positive ratio measured by the number of false alarms per 
frame (FAF), sometimes also referred to as false positives per image 
(FPPI) in the object detection literature.

\begin{figure*}[t]
\centering
\def\svgwidth{1\linewidth}
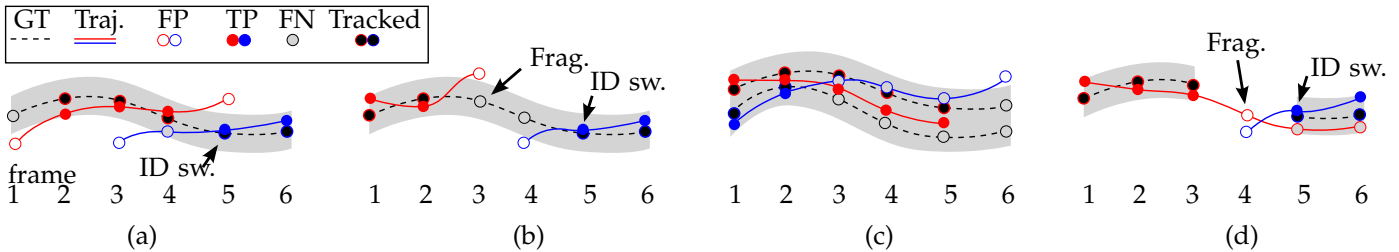
\caption
{
Four cases illustrating tracker-to-target assignments. (a) An ID switch 
occurs when the mapping switches from the previously assigned red track 
to the blue one. (b) A track fragmentation is counted in frame 3 because 
the target is tracked in frames 1-2, then interrupts, and then 
reacquires its `tracked' status at a later point. A new (blue) track hypothesis also 
causes an ID switch at this point. (c) Although the tracking results is 
reasonably good, an optimal single-frame assignment in frame 1 is 
propagated through the sequence, causing 5 missed targets (FN) and 4 
false positives (FP). Note that no fragmentations are counted in frames 
3 and 6 because tracking of those targets is not resumed at a later 
point. (d) A degenerate case illustrating that target re-identification 
is not handled correctly. An interrupted ground truth trajectory will cause a fragmentation. Note the less intuitive ID switch, 
which is counted because blue is the closest target in frame 5 that is 
not in conflict with the mapping in frame 4. 
}
\label{fig:mapping}
\end{figure*}

Obviously, it may happen that the same target is covered by multiple 
outputs. The second prerequisite before computing the numbers is then to 
establish the correspondence between all annotated and hypothesized 
objects under the constraint that a true object should be recovered at 
most once, and that one hypothesis cannot account for more than one 
target. 

For the following, we assume that each ground truth trajectory has one 
unique start and one unique end point, \ie that it is not fragmented. 
Note that the current evaluation procedure does not explicitly handle 
target re-identification. In other words, when a target leaves the 
field-of-view and then reappears, it is treated as an unseen target with 
a new ID. As proposed in \cite{Stiefelhagen:2006:CLE}, the optimal 
matching is found using Munkre's (a.k.a.~Hungarian) algorithm. However, 
dealing with video data, this matching is not performed independently 
for each frame, but rather considering a temporal correspondence.
More precisely, if a ground truth object $i$ is matched to hypothesis 
$j$ at time $t-1$ \emph{and} the distance (or dissimilarity) between $i$ 
and $j$ in frame $t$ is below $\simthresh$, then the correspondence 
between $i$ and $j$ is carried over to frame $t$ even if there exists another
hypothesis that is closer to the actual target. A mismatch error (or 
equivalently an identity switch, IDSW) is counted if a ground truth 
target $i$ is matched to track $j$ and the last known assignment was $k 
\ne j$. Note that this definition of ID switches is more similar to 
\cite{Li:2009:CVPR} and stricter than the original one 
\cite{Stiefelhagen:2006:CLE}. Also note that, while it is certainly 
desirable to  keep the number of ID switches low, their absolute number 
alone is not always expressive to assess the overall performance, but 
should rather be considered in relation to the number of recovered 
targets. The intuition is that a method that finds twice as many 
trajectories will almost certainly produce more identity switches. For 
that reason, we also state the relative number of ID switches, i.e., IDSW/Recall.

These relationships are illustrated in \Fig~\ref{fig:mapping}. For 
simplicity, we plot ground truth trajectories with dashed curves, and 
the tracker output with solid ones, where the color represents a unique 
target ID. The grey areas indicate the matching threshold (see next 
section). Each true target that has been successfully recovered in one 
particular frame is represented with a filled black dot with a stroke 
color corresponding to its matched hypothesis. False positives and false 
negatives are plotted as empty circles. See figure caption for more 
details.

After determining true matches and establishing correspondences it
is now possible to compute the metrics. We do so by concatenating all
test sequences and evaluating on the entire benchmark. This is in
general more meaningful instead of averaging per-sequences figures due to
the large variation in the number of targets.

\subsubsection{Distance measure}
\label{sec:distance-measure}

In the most general case, the relationship between ground truth objects 
and a tracker output is established using bounding boxes on the image 
plane. Similar to object detection \cite{Everingham:2012:VOC}, the 
intersection over union (a.k.a. the Jaccard index) is usually employed 
as the similarity criterion, while the threshold $\simthresh$ is set to 
$0.5$ or $50\%$.

\subsubsection{Target-like annotations}

\begin{figure*}
\centering
 \includegraphics[height=3.6cm]{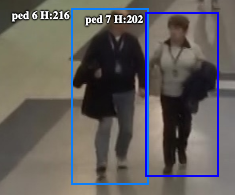}
 \includegraphics[height=3.6cm]{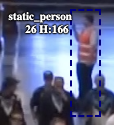}
\includegraphics[height=3.6cm]{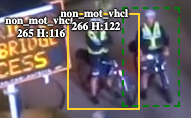}
   \includegraphics[height=3.6cm]{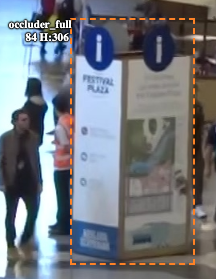}

 \caption{The annotations include different classes. The target class are pedestrians (left). Besides pedestrians there exist special classes in the data such as static person and non-motorized vehicles (non mot vhcl). However, these classes are filter out during evaluation and do not effect the test score. Thirdly, we annotate occluders and crowds.}
 \label{fig:distractors}
\end{figure*}

People are a common object class present in many scenes, but should we track all people in our benchmark?
For example, should we track static people sitting on a bench? Or people on bicycles? How about people behind a glass?  
We define the target class of \CVPR as all upright walking people that are reachable along the viewing ray without a physical obstacle, \ie reflections, people behind a transparent wall or window are excluded.
We also exclude from our target class people on bicycles or other vehicles.
For all these cases where the class is very similar to our target class (see Figure \ref{fig:distractors}), we adopt a similar strategy as in \cite{Mathias:2014:ECCV}. That is, a method is neither penalized nor rewarded for tracking or not tracking those similar classes. 
Since a detector is likely to fire in those cases, we do not want to penalize a tracker with a set of false positives for properly following that set of detections, \ie of a person on a bicycle. Likewise, we do not want to penalize with false negatives a tracker that is based on motion cues and therefore does not track a sitting person.

In order to handle these special cases, we adapt the tracker-to-target assignment algorithm to perform the following steps:

\begin{enumerate}
\item At each frame, all bounding boxes of the result file are matched to the ground truth via the Hungarian algorithm.
\item In contrast to \MOTseven we account for the very crowded scenes and exclude result boxes that overlap $>75\%$ with one of these classes (distractor, static person, reflection, person on vehicle) and remove them from the solution in the detection challenge.
\item During the final evaluation, {\it only} those boxes that are annotated as {\it pedestrians} are used.
\end{enumerate}

\subsubsection{Multiple Object Tracking Accuracy}
\label{sec:mota}
The MOTA \cite{Stiefelhagen:2006:CLE} is perhaps the most widely used 
metric to evaluate a tracker's performance. The main reason for this is 
its expressiveness as it combines three sources of errors defined above:
\begin{equation}
\text{MOTA} = 
1 - \frac
{\sum_t{(\text{FN}_t + \text{FP}_t + \text{IDSW}_t})}
{\sum_t{\text{GT}_t}},
\label{eq:mota}
\end{equation}
where $t$ is the frame index and GT is the number of ground truth 
objects. We report the percentage MOTA $(-\infty, 100]$ in our 
benchmark. Note that MOTA can also be negative in cases where the number 
of errors made by the tracker exceeds the number of all objects in the 
scene.

Even though the MOTA score gives a good indication of the overall 
performance, it is highly debatable whether this number alone can serve 
as a single performance measure. 

{\bf{Robustness.}}
One incentive behind compiling this benchmark was to reduce dataset bias
by keeping the data as diverse as possible. The main motivation is to
challenge state-of-the-art approaches and analyze their performance in
unconstrained environments and on unseen data. Our experience shows that
most methods can be heavily overfitted on one particular dataset, and may not be general enough to handle an entirely
different setting without a major change in parameters or even in the
model.

To indicate the robustness of each tracker across \emph{all} benchmark
sequences, we show the standard deviation of their MOTA score.

\subsubsection{Multiple Object Tracking Precision}
\label{sec:motp}

The Multiple Object Tracking Precision is the average dissimilarity 
between all true positives and their corresponding ground truth targets. 
For bounding box overlap, this is computed as 
\begin{equation}
\text{MOTP} = 
\frac
{\sum_{t,i}{d_{t,i}}}
{\sum_t{c_t}},
\label{eq:motp}
\end{equation}
where $c_t$ denotes the number of matches in frame $t$ and $d_{t,i}$ is 
the bounding box overlap of target $i$ with its assigned ground truth 
object. MOTP thereby gives the average overlap between all correctly 
matched hypotheses and their respective objects and ranges between 
$\simthresh := 50\%$ and $100\%$.

It is important to point out that MOTP is a 
measure of localization precision, \emph{not} to be confused with the 
\emph{positive predictive value} or \emph{relevance} in the context of 
precision / recall curves used, \eg, in object detection.

In practice, it mostly quantifies the localization accuracy of the detector, 
and therefore, it provides little information about the actual performance of the tracker.

\subsubsection{Track quality measures}
\label{sec:track-measures}

Each ground truth trajectory can be classified as mostly tracked (MT), 
partially tracked (PT), and mostly lost (ML). This is done based on how 
much of the trajectory is recovered by the tracking algorithm. A target 
is mostly tracked if it is successfully tracked for at least $80\%$ of 
its life span. Note that it is irrelevant for this measure whether the 
ID remains the same throughout the track. If a track is only recovered 
for less than $20\%$ of its total length, it is said to be mostly lost 
(ML). All other tracks are partially tracked. A higher number of MT and 
few ML is desirable. We report MT and ML as a ratio of mostly tracked 
and mostly lost targets to the total number of ground truth 
trajectories.

In certain situations one might be interested in obtaining long, 
persistent tracks without gaps of untracked periods. To that end, the 
number of track fragmentations (FM) counts how many times a ground truth 
trajectory is interrupted (untracked). In other words, a fragmentation 
is counted each time a trajectory changes its status from tracked to 
untracked and tracking of that same trajectory is resumed at a later 
point. Similarly to the ID switch ratio 
(\cf~\Sec~\ref{sec:tracker-assignment}), we also provide the relative 
number of fragmentations as FM / Recall.

\subsubsection{Tracker ranking}
\label{sec:ranking}
As we have seen in this section, there are a number of reasonable 
performance measures to assess the quality of a tracking system, which 
makes it rather difficult to reduce the evaluation to one single number. 
To nevertheless give an intuition on how each tracker performs compared 
to its competitors, we compute and show the average rank for each one by 
ranking all trackers according to each metric and then averaging across 
all performance measures.

 \section{Conclusion and Future Work}
 \label{sec:conclusion}
 
We have presented a new challenging set of sequences within the \MOTChallenge benchmark. 
Theses sequences contain a large number of targets to be tracked and the scenes are substantially more crowded when compared to previous \MOTChallenge releases. The scenes are carefully chosen and included indoor/ outdoor and day/ night scenarios.

We believe that the \MOTtwenty release within the already established \MOTChallenge benchmark
provides a fairer comparison of state-of-the-art tracking methods, and challenges researchers to develop more generic
methods that perform well in unconstrained
environments and on very crowded scenes.

\ifCLASSOPTIONcaptionsoff
  \newpage
\fi

\bibliographystyle{ieee}

\bibliography{refs-lau,refs-short,refs-anton,refs-new}

\end{document}